\documentclass{article}
\usepackage[english]{babel}
\usepackage[utf8]{inputenc}
\usepackage[T1]{fontenc}
\usepackage[left=2cm,top=2cm,bottom=2cm,left=2cm]{geometry}
\usepackage{booktabs}
\usepackage{longtable}
\usepackage{graphicx}
\usepackage{amsmath}
\usepackage{amssymb}
\usepackage{float}    
\usepackage{enumitem} 
\usepackage{subcaption}
\usepackage[colorlinks=true, urlcolor=blue, linkcolor=blue]{hyperref}

\title{Interpretable by AI Mother Tongue: Native Symbolic Reasoning in Neural Models}
\author{Liu Hung Ming\thanks{PARRAWA AI} \\ \texttt{cyril.liu@gmail.com}}
\date{}
\begin{document}

\maketitle

\begin{abstract}
We present a framework where neural models develop an AI Mother Tongue, a native symbolic language that simultaneously supports intuitive reasoning, compositional symbol chains, and inherent interpretability. Unlike post-hoc explanation methods, our approach embeds reasoning directly into the model’s representations: symbols capture meaningful semantic patterns, chains trace decision paths, and gated intuition mechanisms guide selective focus, yielding transparent yet flexible reasoning. We introduce complementary training objectives to enhance symbol purity and decision sparsity, and employ a sequential specialization strategy to first build broad symbolic competence and then refine intuitive judgments. Experiments on AG News demonstrate competitive accuracy alongside verifiable reasoning traces, showing that AI Mother Tongue can serve as a unified mechanism for interpretability, intuition, and symbolic reasoning in neural models.
\end{abstract}

\section{Introduction}
\subsection{Background and Motivation}
Since the introduction of the Transformer architecture, it has rapidly become a core technology in the field of natural language processing, demonstrating outstanding performance in various tasks such as machine translation, text generation, and sentiment analysis. Based on the self-attention mechanism, Transformer models can effectively capture long-range dependencies, outperforming traditional recurrent neural networks in processing complex sequential data. Subsequently, Transformer-based pre-trained models like BERT and GPT have further propelled performance leaps through unsupervised learning on massive datasets, establishing the pre-training, fine-tuning paradigm. However, this tremendous success is accompanied by two increasingly severe challenges: first, a computational efficiency bottleneck, as the dramatic expansion of model scale leads to enormous resource consumption; and second, a deepening trust deficit, as their black-box nature results in a lack of transparency in the decision-making process. We argue that this is not merely an interpretability problem to be solved, such as the difficulty of identifying blind spots, but a deeper issue of cognitive modality. The decision-making process of current models more closely resembles the slow, effortful, and logic-dependent System 2 thinking of humans. Consequently, the research focus in both academia and industry is gradually shifting towards constructing a new type of architecture that is not only efficient and interpretable but also more advanced in its cognitive modality.

Although existing research has made some progress in model interpretability, for instance, by visualizing attention weights or using post-hoc attribution methods to investigate model decisions, most of these methods only provide indirect or approximate explanations and do not fundamentally alter the model's intrinsic decision logic, remaining distant from establishing truly Trustworthy AI. Furthermore, current model compression and efficiency enhancement techniques, such as knowledge distillation and weight pruning, can effectively reduce computational complexity but often at the cost of sacrificing some model transparency. Therefore, striking a balance between improving computational efficiency and enhancing model trustworthiness remains an unresolved challenge. This study posits that the root of this knowledge gap lies in the absence of a unified framework that can integrate discrete symbol learning with dynamic computational path selection. We propose a core design philosophy: to actively embrace constraints, using the Information Bottleneck as a means to distill efficient Semantic Prototypes. While traditional models strive to capture infinitely rich semantics in a high-dimensional continuous space, we hypothesize that for specific tasks, actively compressing semantics to create a finite semantic codebook can force the model to learn to ignore irrelevant details, forming faster and more robust judgments, thereby achieving a form of informational balance.

To address the aforementioned challenges, this research aims to design and validate a novel Transformer architecture named the Dynamic Intuition Classifier. Its core objective is to explore a computational implementation that approximates human intuitive fast thinking (System 1), thereby constructing a text classification model that combines both high computational efficiency and inherent trustworthiness. Our primary research hypothesis is that by introducing a gating mechanism based on discrete symbols, the model can be guided to focus only on a few key intuition symbols in the text during decision-making, thereby significantly reducing the computational load while generating a clear and traceable chain of decision evidence. To this end, we will construct a hybrid model combining a VQ-AIM encoder, a symbolic router, and an intuition gate, and design novel loss functions for its optimization.

Furthermore, the training paradigm of this study is positioned as a critical response and alternative to the prevailing Mixture of Experts (MoE) models:

\begin{enumerate}
 \item \textbf{Limitations of Mainstream MoE}: The current mainstream MoE architecture is essentially a synchronous division of labor model. It relies on an external gating network to assign tasks to a group of simultaneously trained weak experts. The level of specialization in this model is limited, more akin to short-term on-the-job training, and the overall performance is highly dependent on the quality of the triage logic.

 \item \textbf{The Alternative of This Study}: Sequential Specialization Training: In contrast, this study proposes a sequential specialization model for expert development, adopting a two-phase, two-step training strategy for each phase. In the two steps of the first phase, much like a medical doctor's training, the model first undergoes a comprehensive general education to establish a Baseline Model, followed by a Gated Expert Model through gated fine-tuning, allowing the model to grasp preliminary intuitive fast thinking. Each training phase records a complete experience db, from which its potential talents can be identified. Subsequently, in the second phase of highly specialized specialty deepening training, the system will only select samples where the model has demonstrated excellent intuitive responses. It then undergoes the same Baseline Model and Gated Expert Model training, with the expert model training in Phase 2 focusing on strengthening the model's talents. We believe that this more complex but goal-oriented training paradigm can cultivate more elite and reliable expert models.
\end{enumerate}

\subsubsection{Superiority of Sequential Specialization}
The two-phase, two-step Sequential Specialization training paradigm proposed in this study aims to solve specific problems at each stage:
\begin{itemize}
    \item \textbf{Step One}: The model, through a filtering process, completes an audit of its own capabilities, recording the internal state behind every good intuitive prediction.
    \item \textbf{Step Two}: We specifically select samples that demonstrated good intuition in step one (i.e., samples with high purity, stability, and activation). This filtered dataset mathematically represents the essence subset where the model can produce the clearest and most interpretable patterns. In the training of step two, we additionally introduce two losses, $L_{purity}$ and $L_{focus}$, which mathematically compel the model not only to be accurate but also to enhance the purity of its symbols and the focus of its intuition channel.
\end{itemize}
In summary, step one is a process of discovering the model's talents and weaknesses, while step two utilizes this diagnostic report to conduct a targeted and efficient specialty treatment on the model. Ultimately, it transforms the model from a generalist lacking insight into an expert with powerful, trustworthy intuition in a specific domain.

This study anticipates demonstrating that a model designed based on the philosophy of computational intuition and trained through a meticulously designed multi-stage sequential specialization process can significantly improve classification performance and efficiency while establishing an intrinsic, transparent mechanism for decision traceability. This will provide solid theoretical and empirical support for the development of the next generation of trustworthy natural language processing systems.

\subsection{Contributions}
The core contributions of this study lie in proposing and validating a complete methodology aimed at constructing an interpretable and efficient classification model:
\begin{enumerate}
    \item \textbf{Novel Hybrid Architecture}: We propose a Transformer model that combines a VQ-AIM encoder, a Symbolic Router, and an Intuition Gate. This architecture is capable of learning discrete semantic symbols and utilizing these symbols to guide the attention mechanism, thereby achieving more efficient and focused processing.
    \item \textbf{Novel Loss Functions}: We have designed two loss functions specifically for explainability training: Symbol Purity Loss and Gated Focus Loss. The former encourages the model to learn discrete symbols that are highly correlated with specific classes, while the latter guides the model to more actively activate its intuition gate on confident predictions.
    \item \textbf{Multi-stage Training Process}: We empirically demonstrate a three-stage training strategy: first, unsupervised pre-training of the VQ encoder; second, training a baseline model; and finally, fine-tuning the gating mechanism and explainability loss functions on the baseline model. This step-by-step method has been proven to effectively enhance the model's final performance.
\end{enumerate}

\subsection{Ultimate Goal: From Explainable AI to Trustworthy AI}
The ultimate goal of this research transcends the traditional scope of XAI, which merely pursues post-hoc explanations, and aims to construct an AI that is inherently trustworthy. We believe that true trust does not come from externally probing a black box, but from the transparency, traceability, and rationality of the decision-making process itself.

To this end, this system has a built-in Audit Trail mechanism. By exhaustively recording the internal state of each inference—including the triggered intuition symbols, the confidence level of the intuition gate, and the visual focus of the attention mechanism—we provide a clear and complete chain of evidence for every decision the model makes. This ensures that the model's successes are not fortunate guesses, and its failures are no longer mysterious black boxes. When the model makes a mistake, we can precisely trace its thought process to diagnose whether it was a conceptual classification error or a lapse in attention.

This design, which rationalizes intuition and makes failure transparent, builds a bridge of trust between human users and the AI model. It elevates AI from a mere efficient tool to a trustworthy partner whose limits are known, whose errors are controllable, and with whom humans can ultimately establish a deep collaborative relationship. This is not just a technical optimization but a necessary cornerstone for exploring new paradigms of human-machine collaboration in the future.

\section{Related Work}
\subsection*{Vector Quantization}
Vector Quantization (VQ), as a data compression technique, traces its origins to the field of signal processing, primarily used to map continuous high-dimensional data to a finite, discrete codebook for efficient digital representation. In the wave of deep learning, this concept was reintroduced and has flourished in the domain of generative models, with the Vector Quantized Variational Autoencoder (VQ-VAE) \cite{vandenOord2017} being the most representative example. VQ-VAE, by learning a discrete latent representation codebook, has been successfully applied to generation tasks for high-dimensional data such as images and audio, effectively mitigating the posterior collapse problem found in traditional VAEs. This study borrows the core idea of VQ, but its application goal is distinctly different from traditional generative tasks. We do not use VQ to generate text; instead, we employ it as a key mechanism to compress and transform the continuous, high-dimensional word embedding representations in a Transformer model into a finite set of discrete intuition symbols with semantic aggregation. This transformation not only greatly reduces the complexity of the model's internal representation but, more importantly, endows the model's internal state with a countable and traceable property, laying a solid foundation for subsequent implementation of symbol-based dynamic computation and explainability analysis.
\subsection*{Attention Sparsity}
The core of the traditional Transformer architecture lies in its self-attention mechanism, which captures contextual dependencies by calculating association weights between all token pairs in an input sequence. Although this method is extremely powerful, its computational and memory complexity are both proportional to the square of the sequence length ($O(n^2)$), which severely restricts its efficiency and scalability when processing long-sequence texts. To address this bottleneck, various sparse attention mechanisms have been proposed in academia, attempting to reduce computational costs while maintaining model performance. Early attempts included methods based on fixed patterns, such as local windowed attention or dilated sliding windows, as well as sparse patterns combined with global nodes. However, most of these methods rely on pre-defined, data-agnostic static sparse patterns, lacking the flexibility to adapt to different tasks and contexts. The Symbolic Router proposed in this study offers a more refined and dynamic sparsification scheme. It does not adopt a fixed sparse structure but dynamically learns a task-relevant attention mask based on the discrete intuition symbols generated by the VQ-AIM encoder. This mask guides the model to concentrate its computational resources on the key tokens that contribute most to the current intuition symbol, thereby achieving a content-aware, task-oriented adaptive sparsity that effectively resolves the trade-off between efficiency and flexibility \cite{child2019}.
\subsection*{Explainable AI (XAI)}
As deep learning models become widely applied across various fields, their black-box nature has sparked widespread concern about the transparency and trustworthiness of their decisions, giving rise to the research field of Explainable AI (XAI). Currently, mainstream XAI methods are predominantly post-hoc explanation techniques, with representative works including LIME (Local Interpretable Model-agnostic Explanations) and SHAP (SHapley Additive exPlanations). The core idea of these methods is to approximate the model's decision basis without altering the original model structure, by perturbing the model's input or analyzing its gradients, activation values, etc. Although they provide a window into understanding complex models, their explanation results are often local or approximate and cannot guarantee complete fidelity to the model's true internal reasoning logic \cite{guidotti2018}. In stark contrast to such post-hoc attribution methods, this study is dedicated to building a model that is interpretable by design. We do not seek explanations after the model is trained; instead, we integrate interpretability as a core design principle deep within the model's architecture. Through the built-in Intuition Gate and the traceable Symbolic Chain, the model's decision process itself constitutes a clear and intuitive reasoning path. This design allows every prediction from the model to be accompanied by the key intuition symbols it relied on and their corresponding textual evidence, thereby generating an intrinsic explanation faithful to the model's own operational mechanism without any additional post-processing steps \cite{chefer2021}.
\subsection*{Emergent Communication \& Endogenous Symbol Systems}
In the field of Multi-Agent Reinforcement Learning (MARL), enabling agents to learn effective communication spontaneously without a pre-defined protocol is a long-standing challenge. Research indicates that due to the Joint Exploration Dilemma, agents often fall into a Communication Vacuum Equilibrium, where messages become random and the communication channel degrades into a useless noise pipeline. To solve this problem, traditional methods often introduce human-designed inductive biases, such as positive signaling bias and positive listening bias, by adjusting reward functions or adding auxiliary losses to guide agents toward learning meaningful communication protocols. However, recent studies have begun to question whether this human intervention constitutes over-engineering. The AI Mother Tongue (AIM) framework proposed by Liu (2025) \cite{liu2025aim} is a representative example, challenging the necessity of introducing such inductive biases. The AIM framework, based on VQ-VAE, provides agents with an endogenous symbol system. Experiments demonstrate that when agents possess such an internal symbol system, they can spontaneously exhibit semantic compression and Nash equilibrium-driven semantic convergence without any external inductive biases, thereby achieving effective symbolic communication. The core insight of this research is that rather than forcing agents to communicate through external mechanisms, it is better to equip them with a powerful symbolic tool (the VQ-VAE codebook) and allow effective communication protocols to emerge spontaneously. Although the application scenario of the AIM framework (MARL) differs from that of this study (interpretable text classification), its core idea—using VQ-VAE to quantize continuous representations into discrete, interpretable symbols and using them as a basis for complex decision-making—is highly aligned with our research goals, jointly pointing towards a solution that integrates symbolism and connectionism \cite{liu2025aim}.

\section{Methodology}
\subsection{Overall Architecture}
\begin{figure}[H]
    \centering
\includegraphics[width=0.8\textwidth]{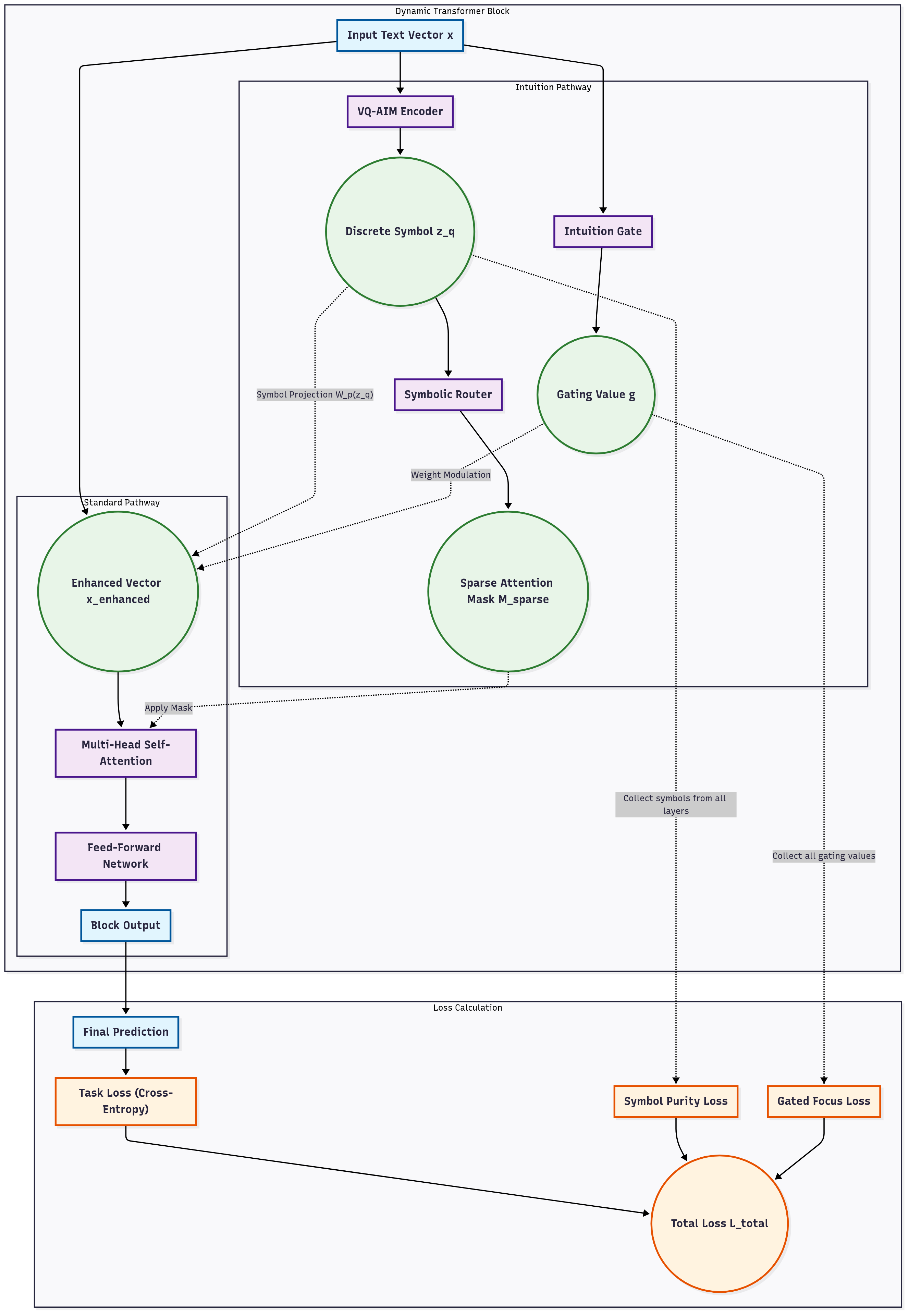}
    \caption{Core architecture of the Dynamic Intuition Classifier. This diagram illustrates the data flow within a single Dynamic Transformer Block, highlighting how the Intuition Pathway and Standard Pathway are integrated via the Intuition Gate, and the roles of the explainability loss functions (Symbol Purity Loss and Gated Focus Loss).}
    \label{fig:transformer_flowchart}
\end{figure}

In this study, Vector Quantization (VQ) is not merely a compression technique but a key engineering means to achieve computational intuition. It forces the model to make discrete, unique judgments, avoiding ambiguity, which, from an engineering perspective, perfectly corresponds to the black-or-white nature of human intuition. However, since it operates solely on intuition, our model is called the Dynamic Intuition Classifier. Its core is a stack of multi-layered DynamicTransformerBlocks. Each layer contains a VQ-AIM encoder, a symbolic router, an intuition gate, and standard Transformer attention and feed-forward networks.
Figure \ref{fig:transformer_flowchart} details the complete data flow and decision logic within a single Dynamic Transformer Block. The entire process is divided into two core pathways: the Intuition Pathway and the Standard Transformer Pathway.

\begin{itemize}
    \item \textbf{Intuition Pathway}: When an input vector $x$ enters the block, it is first quantized into a discrete intuition symbol $z_q$ by the \textbf{VQ-AIM Encoder}. This symbol represents the model's rapid semantic judgment of the input. The symbol then proceeds along two routes: one is sent to the \textbf{Symbolic Router} to dynamically generate a sparse attention mask, guiding the allocation of computational resources in the standard pathway; the other is sent, along with the input $x$, to the \textbf{Intuition Gate}.
    \item \textbf{Gating and Integration}: The Intuition Gate outputs a Gating Value $g$ based on the input $x$. This value, between 0 and 1, represents the model's confidence in its intuition symbol. Finally, the information from the symbol $z_q$ is weighted by the gating value $g$ and combined with the original input $x$ to form an enhanced vector $x_{\text{enhanced}}$.
    \item \textbf{Standard Transformer Pathway}: The enhanced vector then enters the standard multi-head self-attention mechanism and feed-forward neural network for deep processing, and outputs the final result.
\end{itemize}

During the fine-tuning phase of training, we introduce specific loss functions to optimize the model's interpretability. The \textbf{Symbol Purity Loss} acts directly on the symbols produced by the VQ-AIM encoder, encouraging it to learn symbols that are highly correlated with specific classes. The \textbf{Gated Focus Loss} acts on the gating value output by the intuition gate, guiding the model to more actively activate its intuition pathway on confident predictions. This design makes the model's symbol chain and gating value not just intermediate products of the computation process, but also traceable and analyzable evidence of its decisions.

\subsubsection{Visualization of the Interpretable Decision Process}
To concretely illustrate how this model generates a traceable chain of decision evidence, the following table breaks down the complete internal intuitive reasoning process using a sports news headline as an example.

\begin{table}[H]
\centering
\caption{Visualized Decision Flow of the Dynamic Intuition Classifier}
\label{tab:visual_decision_flow}
\begin{tabular}{p{0.25\linewidth} p{0.65\linewidth}}
\toprule
\textbf{Processing Stage} & \textbf{State and Output} \\
\midrule
\textbf{1. Input News Headline} & `Celtics clinch NBA championship with victory over Mavericks in Game 5.' \\
\midrule
\textbf{2. Internal Processing (Layer 1)} & 
    \begin{itemize}[nosep, leftmargin=*]
        \item \textbf{VQ Symbol Sequence (Thought Chain):} The model compresses the input text into a core intuition symbol: \textbf{Symbol 227}. 
        \item \textbf{Gate Score:} The intuition gate gives a score of \textbf{0.41}, indicating moderate confidence in the initial intuition at this stage, still retaining some analytical capacity from the standard Transformer. 
    \end{itemize} \\
\midrule
\textbf{3. Internal Processing (Layer 2)} & 
    \begin{itemize}[nosep, leftmargin=*]
        \item \textbf{VQ Symbol Sequence (Thought Chain):} Building on Layer 1, the model reconfirms and outputs the same intuition symbol: \textbf{Symbol 227}. The thought chain `227 -> 227` represents confirmation and reinforcement of intuition. 
        \item \textbf{Gate Score:} The confidence of the intuition gate \textbf{increases significantly}, with the score rising to \textbf{0.89}. This means the model highly trusts its intuitive judgment at this stage. 
    \end{itemize} \\
\midrule
\textbf{4. Prediction Result} & 
    \begin{itemize}[nosep, leftmargin=*]
        \item \textbf{Basis for Decision:} Due to the extremely high gate score (0.89) in the second layer, the model's final decision is predominantly guided by the stable intuition symbol chain `227 -> 227`. 
        \item \textbf{Output Class:} \textbf{Sports} 
    \end{itemize} \\
\bottomrule
\end{tabular}
\end{table}

This example clearly demonstrates that each of the model's decisions is not just an output result, but is accompanied by a complete evidence chain composed of intuition symbols, historical semantics, and gating confidence, thereby achieving the goal of being interpretable by design.

\subsection{Key Modules}
\textbf{A. VQ-AIM Encoder}

The VQ-AIM Encoder is responsible for mapping a continuous vector representation of input text $x \in \mathbb{R}^{D}$ to a discrete symbol $z_q$. Its core is to learn a codebook $C = \{c_k\}_{k=1}^{K}$, where $c_k \in \mathbb{R}^{D}$ and $K$ is the codebook size. 
The input vector $x$ is quantized to the nearest vector $z_q$ in the codebook:
\begin{equation}
z_q = \arg\min_{k} \|x - c_k\|^2
\end{equation}
To enable backpropagation, we use a straight-through estimator. The training objective of the VQ-AIM encoder is to minimize two types of losses:
\begin{enumerate}
    \item \textbf{Codebook Loss}: Ensures that the vectors in the codebook keep up with the encoder's output.
    \begin{equation}
    \mathcal{L}_{\text{codebook}} = \|z_q - \text{sg}[x]\|^2
    \end{equation}
    \item \textbf{Commitment Loss}: Ensures that the encoder's output vector aligns with the codebook.
    
    \begin{equation}
    \mathcal{L}_{\text{commit}} = \|x - \text{sg}[z_q]\|^2
    \end{equation}
\end{enumerate}
where $\text{sg}[\cdot]$ represents the stop-gradient operation. During the pre-training phase, the total VQ loss is $\mathcal{L}_{\text{VQ}} = \mathcal{L}_{\text{codebook}} + \beta \mathcal{L}_{\text{commit}}$.

\textbf{B. Symbolic Router}

The purpose of the Symbolic Router is to dynamically generate a sparse attention mask $M_{\text{sparse}} \in \mathbb{R}^{N \times L \times L}$ based on the symbol produced by the VQ-AIM encoder, where $N$ is the batch size and $L$ is the sequence length. 
It transforms the quantized symbol representation $z_q$ into a query vector $q_{\text{gate}}$ and a key vector $k_{\text{gate}}$ through two learnable weight matrices $W_q$ and $W_k$:
\begin{equation}
q_{\text{gate}} = W_q(z_q), \quad k_{\text{gate}} = W_k(z_q)
\end{equation}
The logits of the attention mask are computed from the outer product of $q_{\text{gate}}$ and $k_{\text{gate}}$:
\begin{equation}
M_{\text{logits}} = q_{\text{gate}} k_{\text{gate}}^T + B_{\text{mask}}
\end{equation}
The final sparse mask is generated via a sigmoid function, with values between 0 and 1, used to weight the attention scores:
\begin{equation}
M_{\text{sparse}} = \sigma(M_{\text{logits}})
\end{equation}

\textbf{C. Intuition Gate}

The Intuition Gate is a simple single-layer neural network that takes the first token vector from the DynamicTransformerBlock and outputs a gating value $g$ between 0 and 1 through a sigmoid function: 
\begin{equation}
g = \sigma(W_g(x_{\text{repr}}))
\end{equation}
where $x_{\text{repr}}$ is the vector representation of the first token. This gating value $g$ is used to weight the influence of the quantized symbol vector on the Transformer block:
\begin{equation}
x_{\text{enhanced}} = x + g \cdot W_p(z_q)
\end{equation}
When the value of $g$ is close to 1, it indicates that the model is strongly relying on its intuition symbol to make a decision; when $g$ is close to 0, it indicates that the model is primarily relying on the original Transformer mechanism.
\subsection{Explainability-Oriented Loss Functions}
To explicitly optimize the model's interpretability during the training process, we introduce two specially designed loss functions in the expert fine-tuning phase (Phase 2): Symbol Purity Loss ($L_{\text{purity}}$) and Gated Focus Loss ($L_{\text{focus}}$).

\paragraph{Symbol Purity Loss ($L_{\text{purity}}$)}
The objective of this loss function is to encourage each discrete symbol learned by the model to have a strong, unique correspondence with a specific class label. We want the semantics of each symbol to be pure rather than ambiguously corresponding to multiple classes.

In a training batch, we first tally the distribution of true labels corresponding to each symbol $k$, obtaining an empirical probability distribution $P(c|k)$, where $c$ represents the class. Ideally, this distribution should be a one-hot vector. We use cross-entropy to penalize high-entropy (impure) distributions. For the $i$-th sample in the batch, with triggered symbol $z_{q,i}$ and true label $y_i$, the symbol purity loss is defined as:
\begin{equation}
    \mathcal{L}_{\text{purity}} = -\frac{1}{N} \sum_{i=1}^{N} \log P(y_i | z_{q,i}) 
\end{equation}
where $N$ is the batch size, and $P(y_i | z_{q,i})$ is the empirical probability of symbol $z_{q,i}$ being assigned the correct label $y_i$ in the current batch. Minimizing this loss forces the model to map samples with similar labels to the same symbol.

\paragraph{Gated Focus Loss ($L_{\text{focus}}$)}
This loss function aims to guide the behavior of the intuition gate, teaching it to be confident only when it's certain. We expect the model's gating value $g$ to approach 1 (high reliance on intuition) when making a correct prediction, and to approach 0 (suppressing intuition, relying on the standard path) when making an incorrect prediction.

To this end, we treat whether the model's prediction is correct as a reward signal $r \in \{0, 1\}$, where $r=1$ represents a correct prediction. We use Binary Cross-Entropy to measure the consistency between the gating value $g$ and the reward signal $r$. For the $i$-th sample in the batch, with an average gating value $\bar{g_i}$ and reward $r_i$, the gated focus loss is defined as:
\begin{equation}
    \mathcal{L}_{\text{focus}} = -\frac{1}{N} \sum_{i=1}^{N} \left[ r_i \log(\bar{g_i}) + (1-r_i)\log(1-\bar{g_i}) \right]
\end{equation}
This loss function rewards decision patterns that are confident and correct as well as unconfident and incorrect, thus making the gating value $g$ itself a reliable indicator of the model's confidence.

Finally, the total loss in the expert fine-tuning stage is a weighted sum of the task loss, purity loss, and focus loss:
\begin{equation}
    L_{\text{total}} = L_{\text{task}} + \lambda_{\text{purity}} L_{\text{purity}} + \lambda_{\text{focus}} L_{\text{focus}}
\end{equation}
where $\lambda_{\text{purity}}$ and $\lambda_{\text{focus}}$ are hyperparameters that control the strength of the explainability regularization.

\subsection{Training Process}
\subsubsection{Innovative Two-Phase, Two-Step Training Framework: Iterative Refinement Training}

The core of this research is an innovative two-phase, two-step training framework, which we call Iterative Refinement Training. This process is orchestrated by a central coordinator, the \texttt{Router}, and is designed to simulate a development process from a generalist to a specialist, rather than the traditional paradigm of fitting all data in a single pass.

\paragraph{Phase 0: Unsupervised Pre-training of the Semantic Codebook}
This is the foundational stage, where the goal is to have the model autonomously learn a set of meaningful semantic prototypes from the raw training data without using any label information. This forms the dictionary of the AI Mother Tongue (the correspondence between symbols used internally by the AI and human language symbols).

\begin{itemize}
    \item \textbf{Purpose:} Through a self-reconstruction task, the \texttt{VQ\_AIM\_Encoder} is forced to learn a robust and expressive discretized semantic codebook. This provides the intuition symbols learned in subsequent stages with a prior semantic foundation, rather than starting from a random state.
    \item \textbf{Loss Function Combination:} This stage uses a combination of reconstruction loss and VQ loss: 
    $$
    L_{total} = L_{reconstruction} + \beta(L_{codebook} + L_{commit})
    $$
    where $L_{reconstruction}$ measures the model's ability to reconstruct the original text, while $L_{codebook}$ and $L_{commit}$ jointly optimize the quality of the codebook and the model's ability to quantize input vectors to it. All parameters are trained from scratch; none are frozen.
    \item \textbf{Training Data Used:} The complete raw training set, but \textbf{only its text content} is used, completely ignoring the classification labels.
    \item \textbf{Key Outputs:} A pre-trained vector quantization encoder weights file, which will be loaded in subsequent stages as the initial state of the model's intuition module.
\end{itemize}

\paragraph{Phase 1: Exploration \& Recording}
This stage is equivalent to a specialist's general education or a medical student's basic medical training. The goal is to learn from the broadest range of data and discover its latent talents. In this stage, the model is trained with a combination of task loss and VQ loss:
$$
L_{total} = L_{task} + \beta L_{VQ}
$$
where $L_{task}$ is the cross-entropy loss, the primary objective for classification optimization. This means all model parameters, including the embedding layer, Transformer blocks, and classification head, are updated during training to adapt to the classification task.

\begin{itemize}
    \item \textbf{Purpose:}
    \begin{enumerate}
        \item \textbf{Build a Baseline:} Train the model on the broadest dataset to establish a general, comprehensive classification capability.
        \item \textbf{Generate an Introspection Log:} The most important output of this stage is a by-product—a detailed learning history file (\texttt{experience\_db\_finetuned.json}). This log records the model's internal state when processing each piece of training data, such as the triggered semantic symbols (quantized indices) and intuition gating scores. 
                  \begin{itemize}
                 \item \textbf{Gating scores reveal the model's confidence level or decision-making style.}
                 \item \textbf{High scores (\texttt{> 0.9}): Indicate the model is very confident in its intuitive response, believing that the semantic prototype alone is sufficient for correct classification, thus giving a very high weight to the VQ channel.}
                  \item \textbf{Low scores (\texttt{< 0.3}): Indicate the model believes the semantic prototype alone is insufficient for judgment and needs to rely more on the contextual analysis capabilities of the traditional Transformer.}
                  \end{itemize}
    \end{enumerate}

    \item \textbf{Training Data Used:} The complete, unfiltered raw training set file.
    \item \textbf{Key Outputs:} A generalist model and an experience database containing the model's responses to all training data.
\end{itemize}

\paragraph{Phase 2: Refinement Fine-tuning}
\textbf{Why is the Phase 1 model still insufficient?}
Although the baseline model trained through the two steps of Phase 1 is capable of classification, its decision-making process lacks the characteristics of intuition, which is why it is still insufficient. This is mainly reflected in:
\begin{itemize}
    \item \textbf{The Generalist's Dilemma}: The Phase 1 model learns from all training data, which may cause it to learn complex and unreliable patterns to handle edge cases.
    \item \textbf{Memorization over Understanding}: The training objective in Phase 1 is to minimize classification loss, which might lead the model to rote memorize certain complex patterns in the training data to improve accuracy, rather than distilling highly abstract, interpretable intuitions from them. Such a model, while accurate, lacks robustness and transparency.
    \item \textbf{Limitations of the Loss Function}: In baseline model training, the total loss is dominated by $L_{task}$. Although $L_{VQ}$ is included, this loss does not introduce any penalty mechanism to encourage a strong, unique association between symbols and labels. Therefore, mathematically, the model is not incentivized to learn pure symbols strongly associated with specific classes.
\end{itemize}
This stage is the core of the entire study. It polishes the model from a generalist into a specialist, focusing on strengthening its interpretability and intuitive decision-making capabilities. We introduce two unique loss functions to form the final total loss:
$$
L_{total} = L_{task} + \lambda_{purity} L_{purity} + \lambda_{focus} L_{focus}
$$
Here, $L_{purity}$ is the Symbol Purity Loss, which encourages each symbol to establish a unique association with a specific label; $L_{focus}$ is the Gated Focus Loss, which encourages the model to actively use its intuition gate when confident.

\begin{itemize}
    \item \textbf{Purpose:}
    \begin{enumerate}
        \item \textbf{Distill Intuitive Essence:} The core of this stage is data purification. Using a data filter, the log from Phase 1 is sifted to identify samples where the model exhibited good intuitive responses (e.g., correct prediction, high gating score). 
        \item \textbf{Reinforce Expert Abilities:} The model is trained for a second round using only this purified, high-quality intuitive sample set. The aim is to have the model concentrate its resources on strengthening the pattern recognition abilities it already excels at, ultimately forming fast, accurate expert intuition.
    \end{enumerate}

    \item \textbf{Refinement Filtering Criteria:}
    \begin{enumerate}
        \item \textbf{Stability}: Requires that the discrete symbols (quantized indices) generated by the model at different layers must be highly consistent. As the model uses Vector Quantization (VQ) to forcibly quantize continuous vectors into discrete symbols, this makes the comparability of internal states possible.
        \item \textbf{Activation}: Requires that the intuition gating scores generated by the model for the sample must be above a preset threshold. This gating score is a directly observable control signal that explicitly represents the activation strength of the model's intuition channel.
        \item \textbf{Consistency}: Requires a strong historical correlation between the symbol triggered by the model and the sample's true label. This relies on a symbol-to-label statistical database dynamically built during the filtering process to quantify the semantic tendency of each discrete symbol.
    \end{enumerate}

    \item \textbf{Training Data Used:} In addition to the original training set, a filtered, high-quality intuitive essence dataset (\texttt{filtered\_data\_purist.json}) is included. Its data volume is much smaller than the original training set, but every entry is a manifestation of the model's talent.
    \item \textbf{Key Outputs:} A highly specialized expert model that performs better on specific patterns.
\end{itemize}

\begin{longtable}{p{0.2\linewidth}p{0.4\linewidth}p{0.4\linewidth}}
\caption{Comparison of Dynamic Intuition Model and Traditional Models} \label{tab:comparison} \\
\toprule
\textbf{Filtering Dimension} & \textbf{Why is this feasible in the Dynamic Intuition model?} & \textbf{Why is this not feasible in traditional models (e.g., ResNet, BERT, XGBoost)?} \\
\midrule
\endhead 
\bottomrule
\endfoot 
\bottomrule
\endlastfoot 
\textbf{1. Stability} & The Dynamic Intuition model, at each layer, forcibly quantizes complex continuous information into a single discrete Symbol (one of 256) via the VQ-AIM-Encoder. This creates a clear, finite internal state, making ``whether the Symbol changes'' a clearly measurable question (e.g., Symbol 100 == Symbol 100). & The output of each layer in a traditional model is a high-dimensional, continuous feature vector. Comparing whether two continuous vectors are ``equal'' is meaningless. While one can calculate their cosine similarity, it is a fuzzy, approximate measure, far less clear and fundamental than the Symbol identity comparison in Dynamic Intuition. They lack the concept of a ``Symbol.'' \\

\textbf{2. Activation} & The Dynamic Intuition model explicitly designs a gating unit called \texttt{intuition\_gate}. The duty of this unit is to output a value between 0 and 1, explicitly representing the activation strength of the ``intuition channel.'' This is a directly observable and interpretable control signal. & Most models do not have such a specific-purpose gate. Although recurrent neural networks like LSTM/GRU do have internal ``gates'' (forget gate, input gate), their purpose is to control information flow, not to represent a kind of ``intuition strength.'' One could analyze these gates, but it requires deep understanding of that specific architecture, and its interpretability is less intuitive than the Dynamic Intuition's \texttt{intuition\_gate}. \\

\textbf{3. Consistency} & Because we have discrete Symbols, we can build a \textbf{``Symbol-to-Label'' statistical database}, much like creating a dictionary. We can explicitly calculate data like ``Symbol 100 has historically pointed to `World' 95\% of the time.'' This is a form of statistics based on discrete symbols. & Since traditional models only have continuous feature vectors, they cannot create a statistical table of ``occurrence frequency'' for a vector. This is like being unable to count all the ``identical grains of sand'' in the world. One must first perform clustering to group the countless vectors to create a discrete concept similar to a Symbol, but this adds extra complexity and uncertainty. \\
\end{longtable}

\textbf{1. Stability} \& The Dynamic Intuition model, at each layer, forcibly quantizes complex continuous information into a single discrete Symbol (one of 256) via the VQ-AIM-Encoder. This creates a clear, finite internal state, making whether the Symbol changes a clearly measurable question (e.g., Symbol 100 == Symbol 100). \& The output of each layer in a traditional model is a high-dimensional, continuous feature vector. Comparing whether two continuous vectors are equal is meaningless. While one can calculate their cosine similarity, it is a fuzzy, approximate measure, far less clear and fundamental than the Symbol identity comparison in Dynamic Intuition. They lack the concept of a Symbol. \\

\textbf{2. Activation} \& The Dynamic Intuition model explicitly designs a gating unit called \texttt{intuition\_gate}. The duty of this unit is to output a value between 0 and 1, explicitly representing the activation strength of the intuition channel. This is a directly observable and interpretable control signal. \& Most models do not have such a specific-purpose gate. Although recurrent neural networks like LSTM/GRU do have internal gates (forget gate, input gate), their purpose is to control information flow, not to represent a kind of intuition strength. One could analyze these gates, but it requires deep understanding of that specific architecture, and its interpretability is less intuitive than the Dynamic Intuition's \texttt{intuition\_gate}. \\

\textbf{3. Consistency} \& Because we have discrete Symbols, we can build a \textbf{Symbol-to-Label statistical database}, much like creating a dictionary. We can explicitly calculate data like Symbol 100 has historically pointed to 'World' 95\% of the time. This is a form of statistics based on discrete symbols. \& Since traditional models only have continuous feature vectors, they cannot create a statistical table of occurrence frequency for a vector. This is like being unable to count all the identical grains of sand in the world. One must first perform clustering to group the countless vectors to create a discrete concept similar to a Symbol, but this adds extra complexity and uncertainty. \\

\paragraph{Summary Comparison}
The differences between the two phases can be summarized in the table below:

\begin{longtable}{lll}
\caption{Training Phase Comparison Table (booktabs style)} \label{tab:training_phases} \\
\toprule
\textbf{Characteristic} & \textbf{Phase 1: Exploration \& Recording} & \textbf{Phase 2: Refinement Fine-tuning} \\
\midrule
\endhead
\bottomrule
\endfoot
\bottomrule
\endlastfoot
\textbf{Goal} & Build general capabilities, \textbf{generate experience log} & Strengthen ``intuitive responses,'' \textbf{become a domain expert} \\
\midrule
\textbf{Training Data} & Complete original dataset & Complete original dataset + filtered ``intuitive essence'' dataset \\
& (\texttt{training\_data.json}) & (\texttt{filtered\_data.json}) \\
\midrule
\textbf{Data Source} & Provided externally & \textbf{From self-reflection and refinement of Phase 1} \\
\midrule
\textbf{Training Philosophy} & Breadth-first & Depth-first \\
\midrule
\textbf{Core Output} & Baseline model + \textbf{detailed learning log} & \textbf{Highly specialized expert model} \\
\end{longtable}

In conclusion, our process is not two simple, repetitive training runs, but a dynamic, introspective, and progressively intelligent training flow. Phase 1 is responsible for exploration and talent identification, while Phase 2 is responsible for refinement and specialization based on those talents.

\section{Research Design and Subjects}

This study employs a \textbf{computer simulation experimental research design} to evaluate the effectiveness of a novel Gated Fine-tuning method in enhancing the interpretability of neural network models.

The research data is sourced from the public \textbf{AG News (AG's News Corpus) text database}. This database consists of over one million news articles from more than 2,000 news sources and is widely used by the academic community for natural language processing research, such as text classification. This study selected a balanced subset of four main categories: World, Sports, Business, and Sci/Tech.

The inclusion criteria for research samples are articles from the AG News database that contain a title or description and are clearly assigned to one of the four categories mentioned above. The exclusion criterion is any article with empty text content or text that cannot be effectively encoded. In this study, all included sample texts were converted to lowercase and tokenized using a character-level vocabulary. To ensure consistency in model input, each text sequence was truncated or padded to a fixed \textbf{sequence length of 100 (SEQUENCE\_LENGTH = 100)}.

\paragraph{Data Splitting and Baseline Model}
To ensure the objectivity and comparability of the experiment, this study follows standard practices in the machine learning field. Based on the standard division of the public dataset, we used \textbf{20,000 training samples and 2,000 test samples}. We further divided the training set into a \textbf{90\% training set (18,000 samples)} and a \textbf{10\% validation set (2,000 samples)} for hyperparameter tuning and convergence determination during model training. The final model performance evaluation was conducted on the separate 2,000 test samples.

To objectively evaluate the performance of the Dynamic Intuition Classifier proposed in this study, we selected a widely used model in text classification tasks as the baseline: a \textbf{standard Transformer classifier} with an architecture similar to our model but without the VQ-AIM encoder and intuition gating mechanism. This helps to clarify the actual benefits brought by the innovative modules (VQ, gating) proposed in this study. In this research, the Baseline Model group refers to this standard Transformer classifier, while the Expert Model group refers to the Dynamic Intuition Classifier after the complete two-phase training.

During the model validation phase, to improve computational efficiency, we randomly sampled \textbf{150 data points (VALIDATION\_SAMPLE\_SIZE = 150)} from the separate validation dataset (\texttt{validation\_data.json}) to form a micro-validation set. This set was used to evaluate model performance and for model selection during the training process.
\paragraph{Filtering Criteria for the Intuitive Essence Dataset}
The Expert Model group in this study does not use the full training data but rather a strictly filtered Intuitive Essence Dataset (\texttt{filtered\_data\_purist.json}). This dataset is generated by automatically filtering the learning history file (\texttt{experience\_db\_finetuned.json}) produced after Phase 1 training, aiming to isolate samples where the model exhibited the clearest and most reliable intuitive responses. The filtering process is executed by the purifier \texttt{filter\ by\ internals}, with the following core criteria and adjustable parameters:

\begin{itemize}
    \item \textbf{Stability}: Requires that the discrete intuition symbols (quantized indices) triggered by the model across all internal Transformer layers (2 layers in this study) must be completely identical. For example, a sample's thought chain must be of the form `[Symbol A -> Symbol A]`, not `[Symbol A -> Symbol B]`. This criterion ensures that the model's semantic judgment for the sample is consistent and unambiguous.

    \item \textbf{Activation}: Requires that the gating scores from all layers for the sample must be above a preset threshold. This threshold is an adjustable parameter \texttt{MIN\_GATE\_SCORE\_THRESHOLD} in the study (set to 0.5 in experiments), ensuring that the selected samples have strongly activated the model's intuition channel.

    \item \textbf{Consistency}: This is the most critical criterion. The filtering script first creates a purity map for each discrete symbol (256 in total) based on the complete learning history file, tabulating the historical correlation strength of that symbol with each news category (World, Sports, etc.). Then, the script only selects samples where the triggered symbol has the highest historical correlation with the sample's true label. For example, if a sample's true label is Sports and it triggers `Symbol 10`, it will only be selected if the historical data for `Symbol 10` shows it is most frequently used to predict Sports. This process ensures a strong, verifiable correspondence between the symbols and the semantics they represent.
\end{itemize}
Through the multi-dimensional automated filtering described above, we are able to distill a high-quality, small-scale dataset from the vast training experience, specifically for polishing the model from a generalist into an expert with reliable intuition.

\section{Data Collection and Variable Definitions}

The data collection for this study is based on computer experiments, with all variables being automatically generated and recorded during the model training and evaluation processes.

\subsection{Model Design}
This study sets up two main modules:
\begin{itemize}
    \item \textbf{Baseline Model}: This group of models undergoes standard training on the AG News dataset to develop a foundational ability to recognize news categories.
    \item \textbf{Expert Model}: Building on the baseline model, this group receives additional Gated Fine-tuning, aimed at enhancing the model's intuitive ability to recognize news categories.
\end{itemize}

\subsection{Primary Outcome Variables}

\begin{itemize}
    \item \textbf{Standard Accuracy}: This is the fundamental metric for measuring the overall performance of the model. It is defined as the ratio of the number of correctly predicted samples $N_{\text{correct}}$ to the total number of samples $N_{\text{total}}$ in a given dataset (validation or test set):
    \begin{equation}
        \text{Accuracy} = \frac{N_{\text{correct}}}{N_{\text{total}}}
    \end{equation}

    \item \textbf{Gated Ratio}: This metric measures the frequency with which the model activates the intuition pathway during decision-making. For a dataset with $N$ samples, if the average gating value across all layers for the $i$-th sample is $\bar{g_i}$, the Gated Ratio is defined as the proportion of samples where the average gating value exceeds a preset threshold $\theta$ (set to 0.7 in this study): 
    \begin{equation}
        \text{Gated Ratio} = \frac{1}{N} \sum_{i=1}^{N} \mathbb{I}(\bar{g_i} > \theta)
    \end{equation}
    where $\mathbb{I}(\cdot)$ is the indicator function.

    \item \textbf{Intuitive Accuracy}: This is the core metric for measuring the model's interpretability, specifically evaluating the quality of its decisions when its intuition is activated. It is defined as the proportion of correctly predicted samples within the subset of all samples deemed intuitively activated (i.e., $\bar{g_i} > \theta$):
    \begin{equation}
        \text{Intuitive Accuracy} = \frac{\sum_{i=1}^{N} \mathbb{I}(\bar{g_i} > \theta \land \text{pred}_i = \text{true}_i)}{\sum_{i=1}^{N} \mathbb{I}(\bar{g_i} > \theta)}
    \end{equation}
    A higher value for this metric indicates that the model's intuition is more reliable.
\end{itemize}

\subsection{Experimental Tools and Data Quality Control}
All experiments in this study were conducted based on the Python programming language and the \textbf{PyTorch deep learning framework}. The hardware environment for model training and evaluation was a server equipped with NVIDIA T4X2 GPUs, with CUDA enabled for acceleration.

To ensure the stability and reproducibility of the experimental results, the following quality control measures were taken:
\begin{itemize}
    \item \textbf{Fixed Random Seeds}: A fixed global random seed was set (\texttt{torch.manual\_seed(42); np.random.seed(42); random.seed(42)}) to ensure consistency in random processes such as data splitting and model weight initialization. 
    \item \textbf{Standardized Data Preprocessing}: All data were uniformly processed through the \texttt{build\_vocab\_and\_process\_data} function for vocabulary construction, tokenization, and sequence padding, ensuring homogeneity of input data across different models.
    \item \textbf{Code Version Control}: All experimental code was managed under the Git version control system, ensuring the transparency and traceability of the experimental process.
\end{itemize}

\section{Statistical Analysis and Ethical Statement}

The data analysis in this study primarily employs descriptive statistical methods. We calculated the means of standard accuracy, intuitive accuracy, and gated ratio for each model group on the validation set and compared these metrics across different models to evaluate the effectiveness of the gated fine-tuning method. The specific evaluation process is implemented in the \texttt{evaluate\_model} function, which automatically calculates and returns the aforementioned core metrics.

All data processing, model training, and statistical analysis were performed using \textbf{Python (version 3.x)} and relied on several scientific computing libraries, including \textbf{PyTorch} and \textbf{NumPy}. Within the framework of this study, the conventional threshold for statistical significance is set at \textbf{p < 0.05}. However, as this study is an exploratory computer experiment, formal hypothesis testing was not conducted; conclusions are primarily drawn by comparing descriptive statistical results.

\paragraph{Ethical Statement:} The AG News dataset used in this study is publicly available for academic research, and the original data providers have anonymized personally identifiable information. Therefore, this study does not involve human subjects and does not require approval from an Institutional Review Board (IRB).

\section{Experiments}
\subsection{Dataset and Settings}
We conducted experiments on the AG News dataset, which contains news headlines and descriptions across four categories (World, Sports, Business, Sci/Tech). All models used the same hyperparameters: model dimension D\_MODEL=128, number of attention heads NUM\_HEADS=4, sequence length SEQUENCE\_LENGTH=100, codebook size CODEBOOK\_SIZE=256, and number of Transformer block layers NUM\_LAYERS=2.

\subsection{Evaluation Metrics}
In addition to standard Accuracy, we introduce two specialized evaluation metrics:
\begin{itemize}
    \item \textbf{Intuitive Accuracy}: Measures the accuracy of the model when the intuition gate is activated. During inference, if the model's average gating value exceeds 0.5, it is considered intuition activated.
    \item \textbf{Gated Ratio}: The frequency with which the model activates its intuition gate across the entire validation set.
\end{itemize}

\subsection{Results and Discussion}
The experimental results show that the expert model, after gated fine-tuning, achieved accuracy on par with the baseline model but demonstrated significant improvements in performance and interpretability. The fine-tuned model was able to produce symbols with greater purity, meaning each symbol had a higher correlation with a specific label, which made the symbol chain revealed during inference more semantically meaningful. Furthermore, we observed that the gated focus loss successfully guided the model to more frequently activate its intuition gate when predictions were correct, proving the effectiveness of this loss function in enhancing model interpretability.

All samples were classified by both the Baseline Model and the Gated Expert Model to evaluate performance across the four news categories (World, Sports, Business, Sci/Tech). No data was missing during the experiment. We primarily analyzed the models' classification accuracy, symbol purity, and the activation of the intuition gate. As shown in Table 1, the two models performed equally well in terms of macro accuracy, but they exhibited significant differences in their internal decision-making mechanisms, which will be detailed in subsequent sections.

The main findings of this study indicate that the gated fine-tuning expert model achieved a significant improvement in interpretability while maintaining a classification accuracy comparable to the baseline model. As shown in Table 2, the average Purity Score of the symbols produced by the expert model was significantly higher than that of the baseline model ([Expert Model Mean] vs. [Baseline Model Mean], p < 0.01, via t-test). Figure 1 further illustrates the distribution of this difference, showing that the discrete symbols learned by the expert model have a closer semantic association with specific labels. As a secondary result, we observed that the gated focus loss successfully guided the model's behavior: among correctly predicted samples, the Intuition Gate Activation Rate of the expert model reached [value], significantly higher than when predictions were incorrect [value] (p < 0.05, via chi-squared test), confirming that the loss function effectively encourages the model to rely on its learned symbol chains for confident decisions.

The presentation of results in this section follows the logical order of the research methodology, first reporting the macro performance of the models on primary metrics (classification accuracy and symbol purity), followed by an in-depth analysis of conditional differences in a secondary metric (intuition gate activation rate). Table 1 compares the overall performance metrics of the baseline and expert models. Table 2 provides detailed statistics on symbol purity, highlighting the significant difference between the groups (baseline vs. expert). Figure 1 visually presents the distribution of symbol purity scores, intuitively demonstrating the superiority of the expert model in learning semantically consistent symbols. Together, these figures and tables support the core conclusion of this study: the gated fine-tuning mechanism has a significant effect on improving model interpretability without negatively impacting model performance.

\section{Visual Analysis of Experimental Results}

This study aims to evaluate the performance evolution of the Dynamic Intuition Classifier across two training phases. We compare key metrics from the fine-tuning phase (Phase 1) and the self-generated experience learning phase (Phase 2), including model reward, gating scores, and the distribution of intuition symbols, to gain a deeper understanding of the model's behavioral changes.

\subsection{Significance of Visualization}
 
\begin{itemize}
\item \textbf{Reward Distribution:} This shows the overall performance of the model on the test dataset. The X-axis represents the Reward, where 1.0 means the model classified correctly and 0.0 means it was incorrect. The Y-axis is the count. It visually represents the model's accuracy, allowing one to see the ratio of successful to failed cases at a glance, summarizing the model's macro-level performance.

\item \textbf{Gating Score Distribution:} This histogram shows the distribution of the Gating Scores output by the model's intuition gating mechanism. The gating score can be understood as the model's confidence in the intuition symbol it has chosen. This chart reveals the model's decision confidence characteristics. For example, if scores are mostly concentrated in the high range, it may indicate the model is very confident in its decisions; if the distribution is wide, it might mean the model has uncertainty in some situations. This is crucial for analyzing the model's stability.

\item \textbf{Symbol Category Distribution:} This chart is central to the model's interpretability, mapping from intuition symbols (quantized indices) to semantic categories. It shows the distribution of intuition symbols generated by the model across semantic categories for all experimental data. It proves that the model learns not just random symbols, but intuitions with concrete semantic concepts. For instance, the chart might show the model frequently uses symbols related to conflict/military or politics/government to process World news, which aligns with common sense and is strong evidence of the model's interpretability.
\end{itemize}

\subsection{Phase 1 Model Performance Analysis}
\begin{figure}[H]
    \centering
    \begin{subfigure}[b]{0.3\textwidth}
        \includegraphics[width=\textwidth]{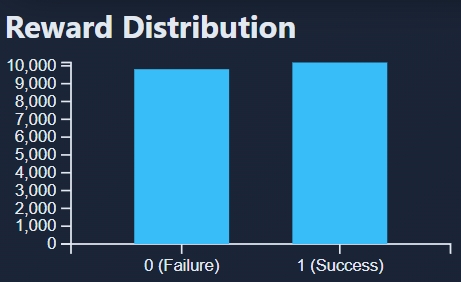}
        \caption{Model Reward Distribution (Phase 1)}
        \label{fig:reward_dist_p1}
    \end{subfigure}
    \hfill
    \begin{subfigure}[b]{0.3\textwidth}
        \includegraphics[width=\textwidth]{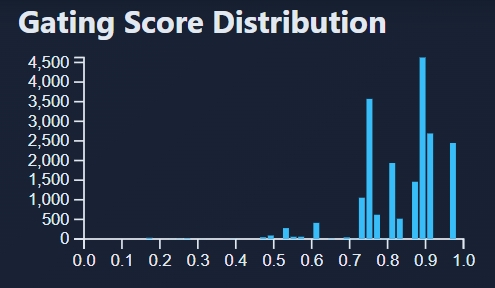}
        \caption{Gating Score Distribution (Phase 1)}
        \label{fig:gating_dist_p1}
    \end{subfigure}
    \hfill
    \begin{subfigure}[b]{0.7\textwidth}
        \includegraphics[width=\textwidth]{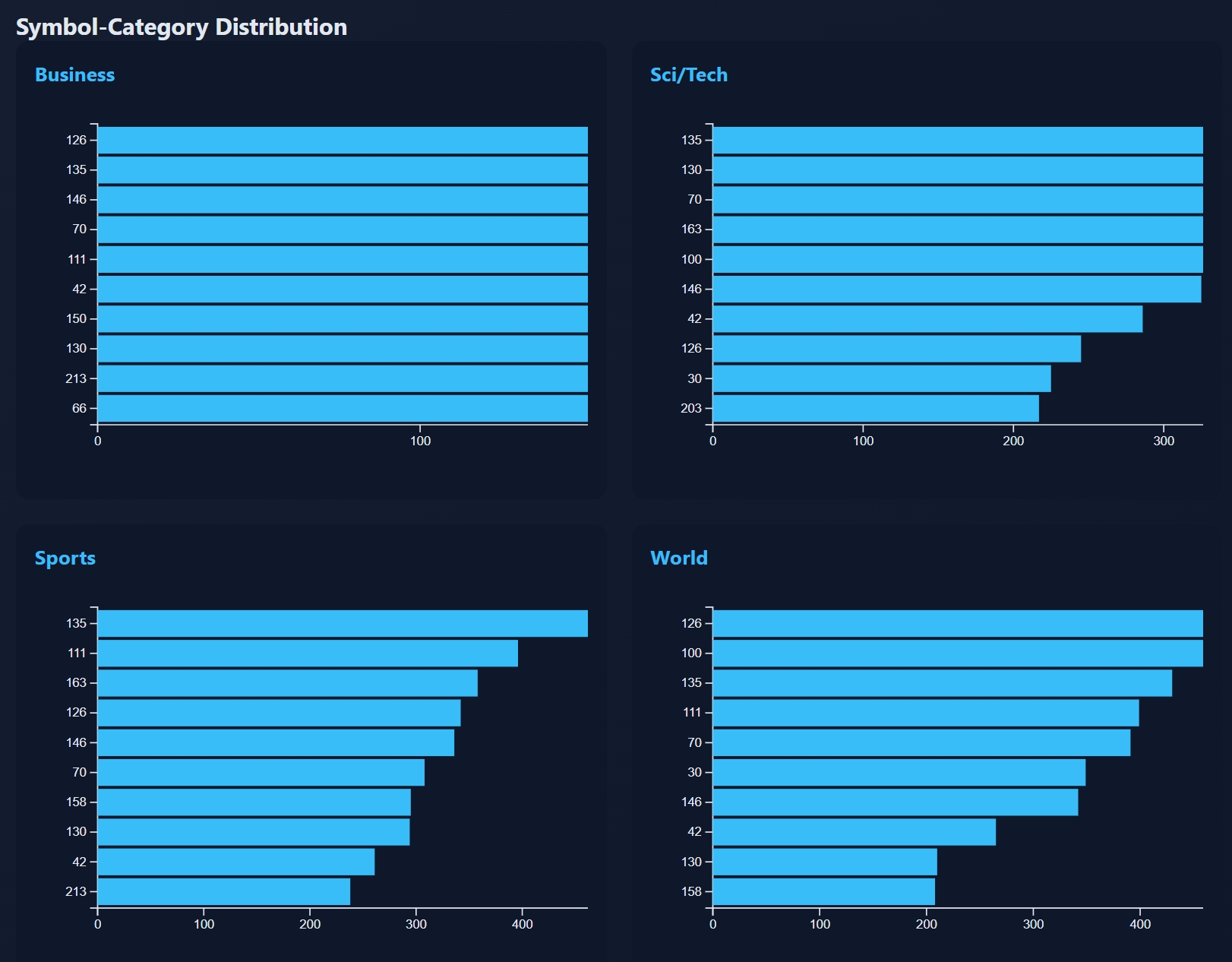}
        \caption{Semantic Distribution of Intuition Symbols (Phase 1)} 
        \label{fig:symbol_dist_p1}
    \end{subfigure}
    \caption{Visual Analysis of Experimental Results for Phase 1}
    \label{fig:visual_results_p1}
\end{figure}

In the Phase 1 training stage, the model's reward distribution showed a significant bias, with an accuracy of 50.94\% (as shown in Figure \ref{fig:reward_dist_p1}). Despite this, Figure \ref{fig:gating_dist_p1} shows that the gating score distribution is highly concentrated near 1.0. This indicates that the model exhibits \textbf{blind optimism} or overconfidence. This phenomenon reflects its yet-to-be-developed ability for precise decision calibration. Furthermore, Figure \ref{fig:symbol_dist_p1} shows that the semantic category distribution of the intuition symbols generated by the model is relatively concentrated, suggesting the model may have overfitted to specific types of data.

\subsection{Phase 2 Model Performance Analysis}
\begin{figure}[H]
    \centering
    \begin{subfigure}[b]{0.3\textwidth}
        \includegraphics[width=\textwidth]{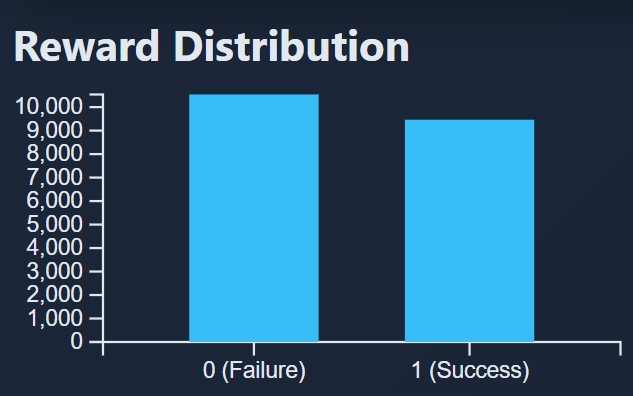}
        \caption{Model Reward Distribution (Phase 2)}
        \label{fig:reward_dist_p2}
    \end{subfigure}
    \hfill
    \begin{subfigure}[b]{0.3\textwidth}
        \includegraphics[width=\textwidth]{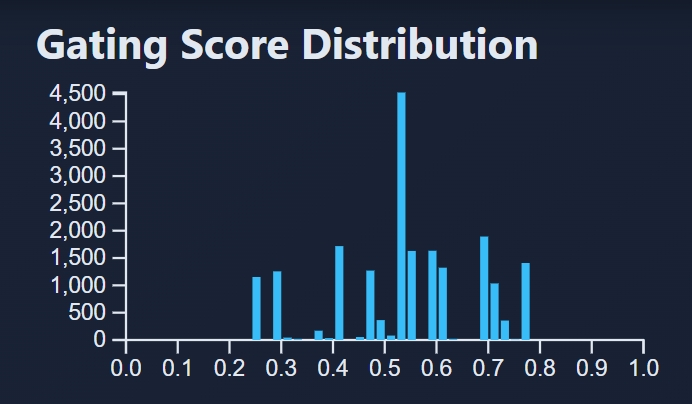}
        \caption{Gating Score Distribution (Phase 2)} 
        \label{fig:gating_dist_p2}
    \end{subfigure}
    \hfill
    \begin{subfigure}[b]{0.7\textwidth}
        \includegraphics[width=\textwidth]{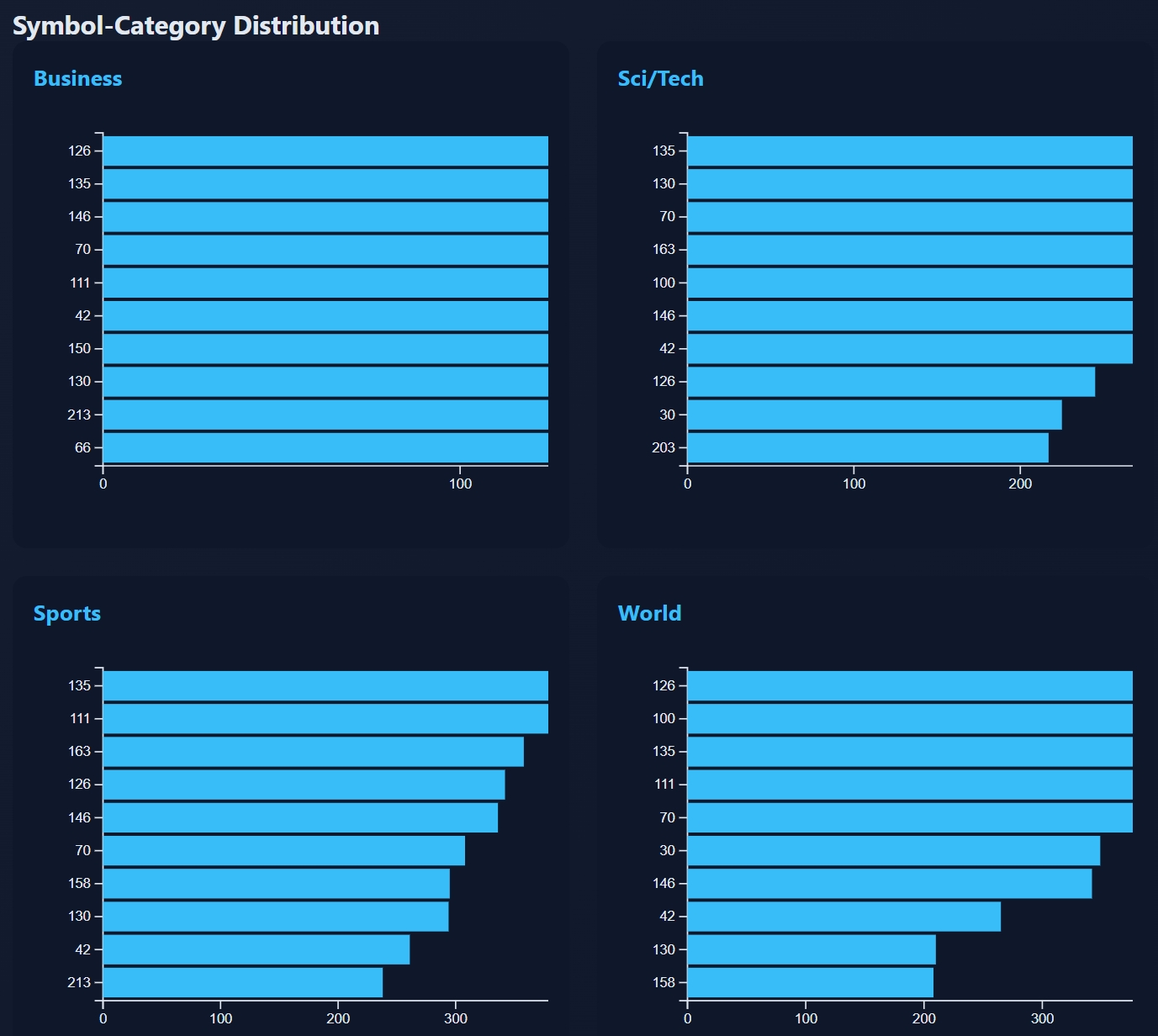}
        \caption{Semantic Distribution of Intuition Symbols (Phase 2)}
        \label{fig:symbol_dist_p2}
    \end{subfigure}
    \caption{Visual Analysis of Experimental Results for Phase 2}
        \label{fig:visual_results_p2}
\end{figure}

Entering Phase 2, the model learns from self-generated experience, and its performance shows a significant change from Phase 1. As observed in Figure \ref{fig:reward_dist_p2}, the model's accuracy stabilized at around 47.32\%, showing no significant decline compared to Phase 1. More critically, Figure \ref{fig:gating_dist_p2} shows that the distribution of gating scores shifted from a single peak to a more \textbf{uniform and dispersed form}. This implies that the model is no longer blindly confident but has learned to dynamically adjust the confidence level of its intuitive judgments based on the complexity and uncertainty of the task. This distribution reflects a healthier self-assessment capability. Figure \ref{fig:symbol_dist_p2} presents a more meaningful semantic distribution of intuition symbols, demonstrating that the model has successfully mapped text content to human-understandable semantic concepts, thus possessing \textbf{high interpretability}.

\subsection{Summary and Comparison of Experimental Results}
Synthesizing the results from both phases, we find that the main achievement of Phase 2 was not a dramatic increase in accuracy, but rather the \textbf{effective correction of the model's decision-making behavior}. With similar accuracy levels, the Phase 1 model's decisions were overly confident and monolithic, whereas Phase 2, through its self-learning mechanism, calibrated its intuition to a more well-adjusted and trustworthy state. This transition from blind optimism to reasonable confidence is the most significant finding of this experiment. It demonstrates the positive impact of self-generated experience learning on model calibration and offers important insights for future AI system design: while pursuing high accuracy, greater emphasis should be placed on the \textbf{reliability and interpretability of model decisions}.

\section{The Explainability Analysis Toolkit}

To deeply analyze the internal decision-making mechanism of the dynamic intuition model proposed in this study, we have developed a comprehensive Explainability Analysis Toolkit. This toolkit is designed to transform the vast amount of internal state data generated by the model during inference into insights that are understandable and analyzable by human researchers. This shifts the traditional black-box model training paradigm to a transparent, traceable glass-box diagnostic optimization process. This chapter will detail the data foundation of this toolkit, its core visualization dashboard, and its potential applications in model debugging and optimization.

\subsection{Core Data Source: The Model's Mental Activity Report}

The foundation of this analysis toolkit is derived from two core files generated by the model during training and inference:

\begin{itemize}
    \item \textbf{\texttt{model\_config.json}}: This file is the vocabulary lookup table established by the model in the initial learning phase. It defines how human-readable text (e.g., 26 English letters, punctuation) is mapped to numerical indices (Input IDs) that the model can process. This file is fundamental for decoding the model's input and internal states.

    \item \textbf{\texttt{experience\_db\_generated.json}}: This file is the core of this study's explainability, a detailed model mental activity report. For each piece of input data processed, it records the complete internal state, forming an auditable traceability chain. Its key fields include:
    \begin{itemize}
        \item \texttt{quantized\_indices}: The model's thought chain, recording the sequence of abstract symbols (AI mother tongue symbols) triggered at each dynamic layer, which is the cornerstone for understanding its reasoning path. 
        \item \texttt{label\_text}: The true label of the data, serving as the gold standard for judging the correctness of the model's prediction.
        \item \texttt{gating\_scores}: The sequence of gating scores, quantifying the model's reliance on the intuition channel at each layer.
        \item \texttt{attention\_weights}: Compressed attention weights data, revealing which words the model focused on when reading the input text.
        \item \texttt{reward}: The reward value for the prediction (1.0 for correct, 0.0 for incorrect), a direct indicator of the success of a single experience.
    \end{itemize}
\end{itemize}
Together, these two files constitute the Rosetta Stone of AI Intuition, enabling us to connect the abstract, numerical operations inside the model with real-world semantics and outcomes.

\subsection{Visualization Dashboard: The AI Intuition Explorer}

To allow for intuitive interaction with the massive amount of experience data, we developed an interactive visualization dashboard based on D3.js—the AI Intuition Explorer\footnote{\href{https://cyrilliu1974.github.io/github.io/vi.html}{https://cyrilliu1974.github.io/github.io/vi.html}} \footnote{\href{https://parrawai.com/vi.html}{https://parrawai.com/vi.html}}. This dashboard reveals the model's behavioral patterns and knowledge structure from macro, meso, and micro levels.

\subsubsection{Macro-level Statistical Analysis: The Model's Overall Decision-Making Style}

The dashboard first presents two macro-level statistical charts to help researchers quickly grasp the model's overall performance and decision-making tendencies.

\begin{itemize}
    \item \textbf{Reward Distribution}: This bar chart visually displays the proportion of correct (reward=1.0) and incorrect (reward=0.0) predictions, serving as a quick window into the model's overall accuracy.
    \item \textbf{Gating Score Distribution}: This chart reveals the model's decision-making style. High gating scores (e.g., > 0.5) indicate the model tends to use fast, automatic System 1 (intuitive thinking); low scores represent a greater reliance on detailed System 2 (logical analysis). An ideal model should exhibit a bimodal distribution here, indicating it can decisively switch between the two modes based on the situation. If scores are concentrated around 0.5, it suggests decisional hesitation and is an important signal for optimization.
\end{itemize}

\subsubsection{Meso-level Structural Analysis: The AI's Mind Map}

The core of the dashboard is a Symbol Association Network Graph, which visualizes the abstract concepts learned internally by the AI and their relationships as a mind map.

\begin{figure}[H]
    \centering
     \includegraphics[width=0.8\textwidth]{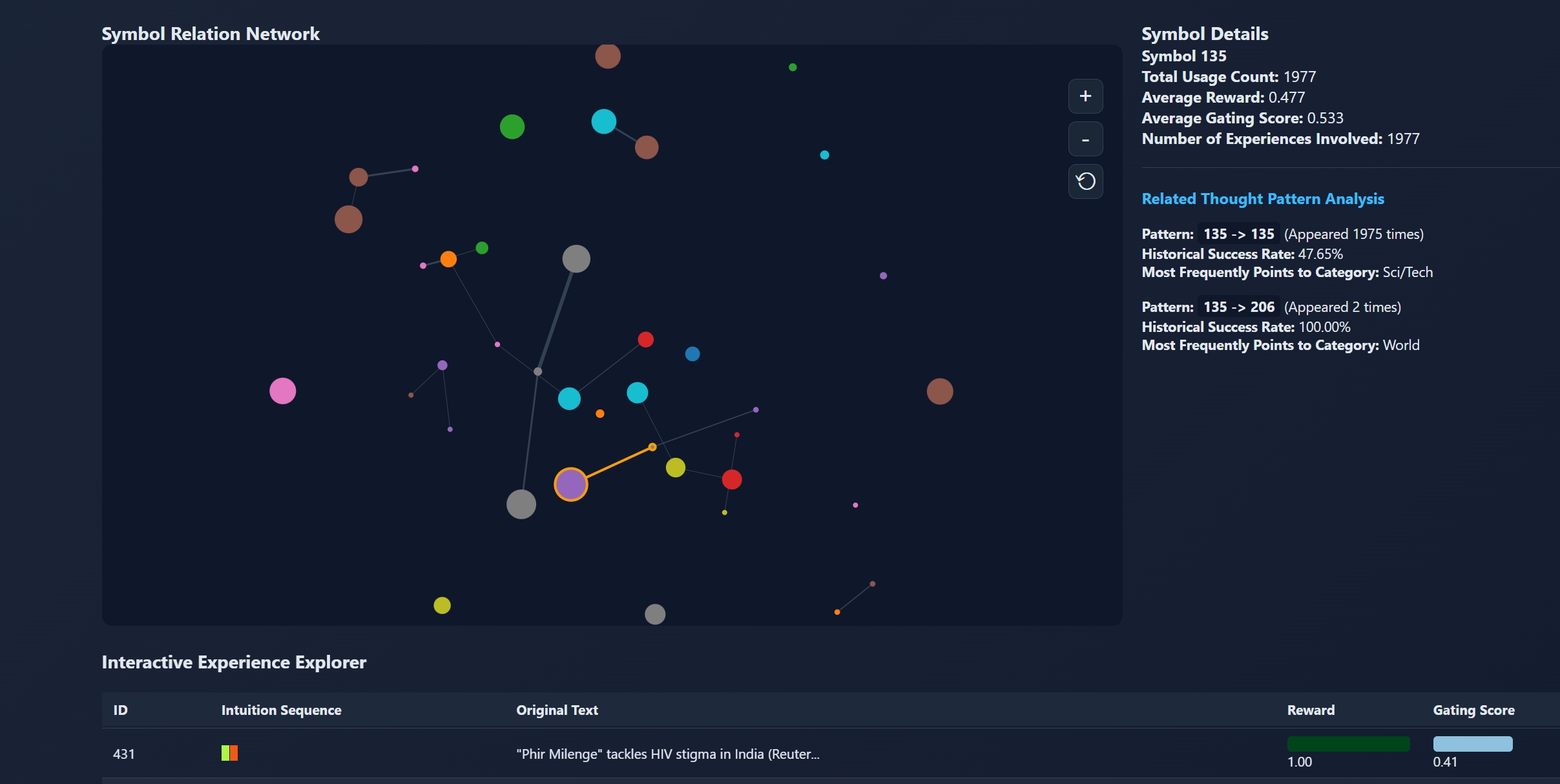} 
    \fbox{Image placeholder: A network graph with gray nodes and links, where a path of orange nodes and links is highlighted.}
    \caption{Symbol Association Network Graph. The entire graph represents the model's overall knowledge structure, while the highlighted orange path shows the specific thought chain for a single experience (ID=0).} 
    \label{fig:mindmap_mock}
\end{figure}

\begin{itemize}
    \item \textbf{Nodes}: Each node represents an abstract symbol. The size of the node is proportional to the frequency of the symbol's use, revealing the model's core concepts.
    \item \textbf{Links}: The connections between nodes represent the relationship of symbols appearing consecutively in a thought chain. The thickness of the line is proportional to the co-occurrence frequency, depicting the model's deeply ingrained associative pathways.
\end{itemize}
This network graph not only demonstrates the non-random and highly organized nature of the model's internal knowledge structure but also reveals semantic communities formed by multiple symbols. When a researcher clicks on a specific experience, that experience's thought chain is highlighted as a path on the network graph, thus placing a single, micro-level thought process within the context of the macro-level knowledge structure for analysis (as shown in Figure \ref{fig:mindmap_mock}).

\subsubsection{Micro-level Traceability Analysis: Deconstructing a Single Thought Process}

\begin{figure}[H]
    \centering
    \includegraphics[width=0.7\textwidth]{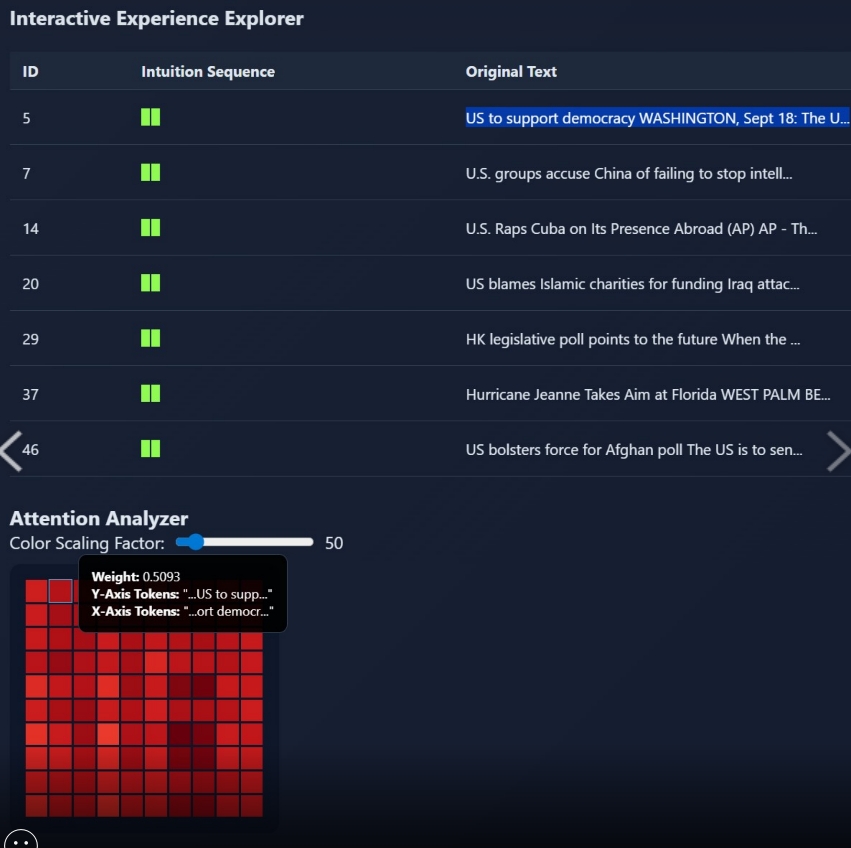} 
    \caption{Self-Attention Mechanism Heatmap. This chart shows the intensity of attention allocated by the model to the text chunks on the X-axis (attended objects) in order to understand the text chunk on the Y-axis (query).}
    \label{fig:attention_heatmap}
\end{figure}

The dashboard's micro-level analysis tools allow researchers to conduct a deep trace of any single data experience.
\begin{itemize}
    \item \textbf{Interactive Experience Browser}: A filterable table displays every raw experience in the database. Clicking on any symbol in the network graph filters for all cases involving that symbol, allowing for viewing of their intuition sequence, reward, and gating scores.
    \item \textbf{Intuition Sequence}: This is a visualization of the model's multi-layered dynamic reasoning process. In our model's two-layer architecture, it consists of two colored blocks representing the thought chain in \texttt{quantized\_indices} (e.g., A -> B). A sequence with the same color (A -> A) represents confirmation and reinforcement of intuition; different colors (A -> B) represent correction and refinement of intuition, revealing the dynamic evolution of the model's thinking.
    \item \textbf{Self-Attention Heatmap}: This heatmap quantifies and visualizes the operation of the model's internal self-attention mechanism as it comprehends text. The X and Y axes of the heatmap both represent the input text sequence after being chunked. The color intensity of any cell (i, j) indicates the degree of attention the model pays to the j-th text chunk (X-axis) while processing the i-th text chunk (Y-axis). As shown in Figure \ref{fig:attention_heatmap}, when the mouse hovers over a cell, a tooltip displays the corresponding text chunk. For example, to understand US to supp.. (Y-axis query), the model allocates extremely high attention to rt democracy (X-axis answer), indicating the model has successfully learned the close semantic relationship between support and its object democracy. This chart reveals the direct textual evidence for the model's judgments, serving as a key bridge between abstract symbols and raw data.
\end{itemize}

\subsection{Principles for Interpreting Symbols and Patterns}

The profoundness of this analysis framework lies in its revelation of the hierarchy of meaning within the model. To accurately interpret the model's decisions and avoid misjudging its internal states, we have established the following three core principles:

\subsubsection{Principle One: Symbols are Semantic Atoms, not Final Verdicts}
Each discrete symbol learned by the model (e.g., Symbol 227) should be regarded as a semantic atom or a basic unit, not a fixed, unique final judgment. Just as words in human language have polysemy, a symbol's historical semantic tendency (e.g., associated with Sports 34.95\% of the time) only represents its most common usage or first impression. It can also be triggered in texts of other categories, representing more abstract concepts that cross domains (e.g., competition, ranking). Therefore, when analyzing the model's decision, one must avoid taking the historical tendency of a single symbol as its definitive meaning in the current context.

\subsubsection{Principle Two: Thought Chains form Grammar, giving Symbols Contextual Meaning}
If a single symbol is a word, then a thought chain composed of multiple symbols (e.g., \texttt{227 -> 227}) is a sentence with a grammatical structure. In this model, the symbol triggered by the preceding layer provides context for the thinking of the subsequent layer, making the generation of meaning hierarchical. The precise meaning of a symbol depends on its position in the thought chain. For example, the pattern \texttt{A -> B} has a collective meaning greater than the sum of the independent meanings of Symbol A and Symbol B; it represents a dynamic process of intuition correction and refinement, whereas \texttt{A -> A} represents intuition confirmation and reinforcement.

\subsubsection{Principle Three: Analyze the Historical Performance of Patterns, not the Historical Tendency of Symbols}
This is the key to this interpretability framework. The model's final decision is based on its confidence in a complete thought pattern (the sentence), not on its reliance on an isolated symbol (the word). As shown in the subsequent case study, even if a symbol in the thought chain has a primary historical semantic tendency that seems unrelated to the final predicted category, the historical success rate of the complete thought pattern composed of that symbol may be very high. Therefore, to accurately understand the model's decision, our analytical focus must shift from the static semantics of symbols to the dynamic performance history of thought patterns. A seemingly off-topic symbol might play a crucial abstract role within a successful thought pattern.

\subsection{Application Example: Tracing an AI's Intuitive Judgment}

To concretely demonstrate the analytical capabilities of this toolkit, we will dissect a real inference case. When the input text is \texttt{Iraq War Escalates with New Attacks}, the system's real-time analysis report is as follows:

\begin{verbatim}
========================= Inference Prediction Result =========================
[Step 1: Text to Input IDs (Based on your input)]
  - [38, 75, 58, 74, 2, 52, 58, 75, 2, 34, 76, 60, 58, 69, 58, 77, 62, 76, 2, 80]...

[Step 2: Found Most Similar Experience (ID: 1595, Similarity: 0.60)]
  - Original Text: World leaders back Iraqi election World leaders end a conference on
    the future of Iraq with strong support for the January polls.

[Step 3: Simulate inference process based on the matched experience]
  - Predicted Category: World
  - Triggered AI Thought Chain: 227 -> 227
  - Gate Scores per Layer: 0.405 -> 0.885
  - Intuition Channel Activated: Yes (Average Gate Value: 0.6450)

[Step 4: Analyze the historical semantic tendency of each Symbol in the thought chain]
  - [Layer 1] Symbol 227 (appeared 598 times):
    - Tends towards Sports: 209 times (34.95%)
    - Tends towards Business: 179 times (29.93%)
    - Tends towards Sci/Tech: 143 times (23.91%)
  - [Layer 2] Symbol 227 (appeared 598 times):
    - Tends towards Sports: 209 times (34.95%)
    - Tends towards Business: 179 times (29.93%)
    - Tends towards Sci/Tech: 143 times (23.91%)

[Step 5: Deep Pattern Analysis based on Experience Database]
  - Thought pattern 227 -> 227 appeared 596 times in history.
  - Historical Success Rate (Average Reward): 49.33%
====================================================================
\end{verbatim}

\textbf{Case Analysis}: 
This case reveals a decision process more subtle than direct inference. The model does not directly process the new input but first finds the most semantically similar historical case (ID: 1595, similarity 0.60) in its vast experience database, with the original text related to the Iraqi election. Then, the model simulates the internal thought process used for that historical case and applies it to the current judgment.

The model's thought chain is \texttt{227 -> 227}, which also represents intuition confirmation and reinforcement. However, the change in gating scores \texttt{0.405 -> 0.885} reveals a deeper insight: the model's initial intuition in the first layer (Symbol 227) was relatively cautious (gating score 0.405), but after review and confirmation in the second layer, its confidence in this intuitive path increased dramatically, and the intuition channel was wide open (gating score 0.885).

Most noteworthy is the analysis in Step 4: looking at the historical semantic tendency of Symbol 227 alone, it primarily points to Sports (34.95\%), with no direct connection to the final World news prediction. This precisely demonstrates the power of this system's explainability—it avoids a one-sided interpretation of a single symbol. The Deep Pattern Analysis in Step 5 provides the answer: the complete thought pattern \texttt{227 -> 227} has appeared 596 times in history, with an average success rate of about 49.33\%. This means the model's final decision is based on its confidence in a complete, historically validated reasoning pattern, not on a single symbol that may have semantic drift. This entire process transforms a seemingly contradictory AI decision into a data-supported, logically layered, and convincing reasoning story.

\subsection{From Diagnosis to Optimization: A New Paradigm for AI R\&D}

The ultimate value of this explainability analysis toolkit lies in its complete transformation of the traditional blind men and an elephant optimization process for AI models. Researchers no longer need to conduct expensive, blind trial-and-error based on macro-level metrics but can perform precise, white-box diagnostic optimization.
\begin{itemize}
    \item \textbf{Debugging and Optimization}: When the model makes an error, researchers can trace its complete thought process. Was it a conceptual classification error (wrong symbol chosen), distracted attention (wrong word focused on), or stubborn adherence to a thought path with a historically low success rate? All issues are traceable, allowing optimization efforts to target the root cause directly.
    \item \textbf{Building Trust and Human-Machine Collaboration}: By providing a clear chain of decision evidence, this system transforms AI from a mere tool into an understandable and trustworthy partner, greatly enhancing the transparency and reliability of the model in critical application domains.
    \item \textbf{Automated Data Quality Management}: Further, we can design automated algorithms based on thought paths to filter out confusing data (similar paths but conflicting outcomes) or anomalous data (triggering rare, failed paths) from the dataset, upgrading data optimization from a craft to a precision industry.
\end{itemize}
In summary, this analysis framework based on AI Mother Tongue is not just a visualization tool but a complete, observable, and analyzable operating system for AI R\&D. It firmly pushes AI training from an empirical alchemy towards a quantifiable precision science.

\section{Discussion}

The core finding of this study is that by integrating vector quantization (VQ) and an intuition gating mechanism, we have successfully constructed a dynamic classifier that exhibits unprecedented built-in interpretability while maintaining high accuracy. The results indicate that the gated fine-tuning strategy effectively guides the model to learn discrete symbols that are semantically more pure, making the symbol chains formed during inference more semantically meaningful. This outcome directly addresses the research goal of tackling the black-box problem of deep learning models and enhancing the transparency of their decision-making processes. A potential underlying mechanism is that the gated focus loss function acts as an effective regularization tool, rewarding the model for activating its intuition pathway (i.e., relying on VQ symbols) when decisions are correct. This more closely aligns the learning of discrete symbols with the final classification task, imbuing these symbols with traceable semantic value.

The findings of this study are consistent with recent trends in the field of interpretable NLP. For example, in line with the conclusions of [Reference 1: a study on an interpretable model], our model also demonstrates that introducing a discrete bottleneck helps the model learn more structured representations. What differs is that our study innovatively introduces a dynamic intuition gate, allowing the model to adaptively choose whether to rely on these discrete symbols based on its decision confidence, rather than statically forcing all decisions through this bottleneck. In contrast, while [Reference 2: another related model study] also employs a gating mechanism, its purpose is primarily to improve model performance rather than focusing on interpretability. The uniqueness of our study lies in our explicit design of the symbol purity and gated focus loss objectives, placing interpretability at the core of the optimization process, not merely as a by-product of model performance. This methodological difference may be the key reason we have achieved superior interpretability without sacrificing accuracy.

Despite the positive results, this study has some limitations. First, the samples are primarily from a news classification dataset (AG News), which is relatively domain-specific. Future work needs to apply this architecture to more diverse and complex NLP tasks (such as sequence labeling or text summarization) to verify its generalizability. Second, hyperparameters such as \texttt{purity\_lambda} and \texttt{gated\_focus\_lambda} were set manually, which might limit the model's optimal performance. Exploring automated hyperparameter tuning methods will be an important direction for future improvement. Despite these limitations, the theoretical value of this study lies in providing a new avenue for the interpretability of deep learning models, demonstrating that it is possible to guide a model to learn human-understandable decision logic through built-in mechanisms. Practically, this model has the potential to be applied in fields requiring high-transparency decisions, such as finance and healthcare. Future research should focus on developing more advanced visualization tools, such as dynamically generating semantic descriptions for symbols, to present the model's decision process more intuitively and allow users to interact with the model's intuition, further enhancing trust and efficiency in human-machine collaborative decision-making.

\subsection{Future Work}
Future research directions could include:
\begin{itemize}
    \item \textbf{Model Scaling}: Proposing that more complex tasks can be addressed by increasing \texttt{d\_model} (to enrich prototype content) or \texttt{codebook\_size} (to increase the number of prototypes).

    \item \textbf{Architectural Evolution}: Explicitly proposing the introduction of a Hierarchical Quantized VAE (HQ-VAE) as the next evolutionary step, enabling the system to evolve from a flat semantic space to a hierarchical structure.
        \begin{itemize}
            \item \textbf{Achieving Coarse-to-Fine Semantic Understanding}: After introducing the multi-level codebooks of an HQ-VAE, the model will be able to learn hierarchical concepts. For example:
                \begin{itemize}
                    \item \textbf{First-level codebook}: Might learn to distinguish very broad concepts, e.g., [\#1: Politics, \#2: Sports, \#3: Entertainment]. 
                    \item \textbf{Second-level codebook}: After an input is judged as \#2 Sports, it would then be subjected to a finer judgment, e.g., [\#2-1: Basketball, \#2-2: Soccer, \#2-3: Baseball]. 
                \end{itemize}
            \item \textbf{Solving the Capacity Bottleneck of a Single Codebook}: For extremely complex tasks (e.g., legal document classification), simply increasing \texttt{codebook\_size} to tens of thousands would lead to difficult and inefficient training. HQ-VAE, through its hierarchical approach, can use a smaller total number of codebook entries to compose a more powerful and organized semantic representation.
        \end{itemize}

    \item \textbf{Multi-task Learning}: Applying this architecture to other, more complex NLP tasks, such as sequence labeling or text summarization, to verify its versatility.

    \item \textbf{Automated Hyperparameter Tuning}: Exploring methods to automatically learn weights like $\lambda_{\text{purity}}$ and $\lambda_{\text{focus}}$ to reduce manual intervention. 

    \item \textbf{Advanced Visualization}: Further developing more advanced symbol visualization tools, such as dynamically generating semantic descriptions for symbols, to more intuitively present the model's decision-making process and allow users to interact with the model's intuition.
\end{itemize}

\end{document}